\documentclass{article}

\PassOptionsToPackage{numbers,sort&compress}{natbib}

\usepackage[dblblindworkshop, final]{neurips_2025}

\usepackage{enumitem}
\usepackage{ragged2e} 
\usepackage{amsmath,amssymb,amsfonts}
\usepackage{algorithmic}
\usepackage{graphicx}
\usepackage{textcomp}
\usepackage{xcolor}
\usepackage{cite}
\usepackage{multirow}
\usepackage{lscape}
\usepackage{adjustbox}
\usepackage{tablefootnote}
\usepackage{tabulary}
\usepackage{array}
\usepackage{tabularray}
\usepackage{float}
\usepackage{booktabs}
\usepackage{subcaption}
\usepackage{tikz}
\usepackage{graphicx}
\usepackage{balance}
\usepackage{expl3}
\usepackage{xcolor}
\usepackage{colortbl}
\usepackage{tabularx}
\usepackage{makecell}
\usepackage[utf8]{inputenc} 
\usepackage[T1]{fontenc}    
\usepackage{hyperref}       
\usepackage{url}            
\usepackage{nicefrac}       
\usepackage{microtype}      

\title{Offline Policy Evaluation of Multi-Turn LLM Health Coaching with Real Users}
\workshoptitle{Multi-Turn Interactions in Large Language Models}

\author{%
  Melik Ozolcer \quad
  Sang Won Bae \\[1ex]
  \texttt{\{mozolcer, sbae4\}@stevens.edu} \\[1ex]
  Stevens Institute of Technology
}

\begin{document}

\maketitle

\begin{abstract}
We study a web-deployed, tool-augmented LLM health coach with real users. In a pilot with seven users (280 rated turns), offline policy evaluation (OPE) over factorized decision heads (\textsc{Tool}/\textsc{Style}) shows that a uniform heavy-tool policy raises average value on logs but harms specific subgroups, most notably low-health-literacy/high-self-efficacy users. A lightweight simulator with hidden archetypes further shows that adding a small early information-gain bonus reliably shortens trait identification and improves goal success and pass@3. Together, these early findings indicate an evaluation-first path to personalization: freeze the generator, learn subgroup-aware decision heads on typed rewards (objective tool outcomes and satisfaction), and always report per-archetype metrics to surface subgroup harms that averages obscure.
\end{abstract}

\section{Introduction}
Wearable devices generate personal health data that AI health coaches powered by LLMs can analyze in real time. These agents can help users understand trends, assess risks, and achieve health goals. However, successful deployment requires more than accurate analysis: agents must maintain trust across multi-turn dialogues, manage long conversations, personalize to individual health profiles, and handle sensitive data competently. Current research relies mainly on synthetic benchmarks, leaving gaps in understanding real-world performance with actual users. 

This paper presents early findings from a web-deployed LLM health agent and outlines an operational framework for personalization. We (i) perform offline policy evaluation (OPE) on deployment logs to compare decision policies (e.g., when to use tools, response style), and (ii) run a lightweight simulation with hidden user archetypes to test whether brief early exploration improves outcomes. Analysis of pilot interactions shows an early period of high-rated, tool-driven engagement that degrades as conversation length increases. While baseline health literacy and self-efficacy are associated with user experience, tool usage and response characteristics emerge as the primary drivers of rating variation. Both OPE and simulation reveal heterogeneous effects across archetypes.

We model health coaching as a user-conditioned partially observable Markov decision process (POMDP) with typed rewards. The per-turn utility is a user-specific composition defined in Eq.~\ref{eq:reward_comp}, where $z_t$ summarizes dialogue context, recent outcomes, and user characteristics (including health literacy and self-efficacy), and the tool component derives from success/failure outcomes. To address cold-start and sparse feedback, we add a small, early-turn user-model information-gain bonus (curiosity) that encourages actions which reduce uncertainty over latent interaction preferences (e.g., tolerance for explanation, desire for analysis) \citep[][]{wan2025enhancing,schmidhuber1991possibility,oudeyer2007intrinsic,pathak2017curiosity}. We define this bonus formally in Eq.~\ref{eq:curiosity} and apply it only in the first few turns.

First, OPE with self‑normalized IPS (SNIPS) and AIPW (doubly robust) \citep[][]{swaminathan2015counterfactual,dudik2011doubly,thomas2016data} lets us compare counterfactual policies on real logs without additional user trials. We find that uniform, heavy tool usage can maximize average reward on logs but causes harm for specific archetypes (e.g., low-literacy/high-efficacy), strengthening the case for archetype-aware reward weighting. Second, in simulation with hidden archetypes, brief early information-gain phase reliably shortens trait-identification time and improves task success and reliability (pass@3), which supports a "probe-then-personalize" strategy consistent with our deployment findings.

\section{Related Work}
\paragraph{Health LLMs and tool-using agents.}
LLM-based agents have advanced health applications in coaching and diagnostics. PH-LLM \citep{cosentino2024towards} fine-tunes models on wearable data for personalized sleep and fitness insights, while PHIA \citep{merrill2024transforming} uses iterative reasoning with tools for health data interpretation across multiple turns. Recent work explores behavior-theoretic scaffolding \citep{jorke2025gptcoach} and proactive strategies \citep{srinivas2025substance} for long-term coaching. However, multi-turn deployment remains challenging: \citep{laban2025llms} report substantial degradation over extended conversations, and tool-enabled settings can be brittle. Notably, \citep{wang2023mint} show that while tools and natural-language feedback generally help performance, multi-turn pipelines remain fragile and certain training choices can hurt longer conversations. 

\paragraph{Personalization, curiosity, and reward modeling.}
Personalization through RL and preference learning shows promise for diverse users. Curiosity and intrinsic-motivation approaches use information gain to reduce uncertainty about user traits. These have been adapted to multi-turn dialogue, education, and fitness, where early exploration can accelerate downstream performance \citep{wan2025enhancing}. Related work includes variational preference learning with latent user variables \citep{poddar2024personalizing}, turn-level credit assignment for tool-using agents \citep{zeng2025reinforcing}, and low-rank reward modeling for efficient personalization \citep{bose2025lore}. We build on these ideas but focus on health coaching with tool use outcomes and survey-informed archetypes (health literacy, self-efficacy). We combine an early information-gain bonus with user-conditioned reward composition.

\paragraph{Evaluation: OPE, simulators, and long-horizon benchmarks.}
Off-policy evaluation (OPE) lets us compare counterfactual policies from logs without needing new user trials, using importance-weighted and doubly-robust estimators with calibration diagnostics \citep{swaminathan2015counterfactual,dudik2011doubly,thomas2016data}. Tool-using agent benchmarks now use multi-turn tasks with typed end-state checks and pass@3 reliability \citep{wang2023mint,yao2024tau,zhou2025sweet}, while long-horizon assistants expose planning and consistency limitations \citep{mialon2023gaia,xie2024travelplanner}. We align with this trend by (i) applying OPE on deployment logs to compare decision policies under selection correction and (ii) using a lightweight simulator with hidden archetypes to test whether early information-gain phases improve trait identification, goal success, and reliability.

\section{Methods}

\subsection{System and Data}
We built and deployed a web-based LLM health system (Qwen3-235B-A22B via API providers \citep{qwen3technicalreport}) for multi-turn health coaching with real users and wearable data. The system processes user-uploaded health data (i.e., Apple Health app export) through: (1) a data pipeline that preprocesses historical data into daily features (sleep, HRV, VO$_2$max, activity) and generates ML predictions for stress, soreness, and injury risk (rolling-origin cross-validation; $R^2{=}\,0.50,\,0.28,\,0.40$ respectively on 1{,}280 day-samples, $N{=}28$) \citep{ozolcer2025sepa}; (2) agent tools including \emph{Code Executor}, \emph{Web Searcher}, and \emph{Email} (reminders and events).

\paragraph{Participants and Procedure.}
Seven students ($N{=}7$) completed a pilot study involving baseline surveys (eHEALS for health literacy \citep{norman2006eheals} and the General Self-Efficacy Scale \citep{schwarzer1995generalized}), followed by multi-turn dialogues over multiple days (typically $>$1 hour, $30{+}$ messages) with per-response 1-5 star ratings. We logged system metrics (latency, token counts, tool usage/outcomes) and client-side interaction proxies (typing duration, edits/pauses/tab-switches). The study received IRB approval and participants provided informed consent. 

\subsection{Tool-Outcome Rubric}
We labeled each tool invocation using a rubric that drives the objective component of reward $R_{\text{tool}}$:
\begin{table}[h]
\centering
\caption{Typed tool-outcome rubric and mapping to $R_{\text{tool}}$. We define $R_{\text{tool}}$ on every turn; when no tool is invoked, $R_{\text{tool}}{=}0$.}
\label{tab:tool_rubric}
\begin{tabular}{l l l c}
\toprule
Tool & Outcome & Criteria & $R_{\text{tool}}$ \\
\midrule
No Tool        & —              & no tool invoked                         & $0$ \\
\addlinespace
Web Searcher   & \textbf{Success} & $\geq$1 URL; claim-evidence consistent   & $+1$ \\
               & \textbf{Failure} & no URL or dead/contradictory link        & $-1$ \\
\addlinespace
Code Executor  & \textbf{Success} & executes; plot/table present; axes/units labeled & $+1$ \\
               & \textbf{Failure} & exception; wrong data slice; no visualization    & $-1$ \\
\addlinespace
Email & \textbf{Success} & event/message created; ISO8601 date/time present & $+1$ \\
               & \textbf{Failure} & not created / confirmation missing               & $-1$ \\
\bottomrule
\end{tabular}

\vspace{0.25em}
\small \emph{Note:} $R_{\text{eng}}$ is a small, bounded engagement signal defined on \emph{all} turns (e.g., mild latency penalty, structure/clarity bonus, evidence gating). The OPE objective uses $R_{\text{obj}}=R_{\text{tool}}+R_{\text{eng}}$ and is reported via SNIPS.
\end{table}

\subsection{Problem Formulation}
We model multi-turn coaching as a user-conditioned partially observable Markov decision process (POMDP). At turn $t$, the agent receives observation $o_t$ (user latest tool outputs) and maintains a belief state $z_t = f_\phi(h_t, u_i, m_t)$ summarizing conversation history $h_t$, user features $u_i$ (eHEALS/GSE strata and derived interaction features), and current metrics $m_t$ (health data, model predictions, recent tool outcomes). The action $a_t$ factorizes into discrete heads:
\[
a_t = (\underbrace{\textsc{Tool}}_{\in\{\varnothing,\ \text{Search},\ \text{Code},\ \text{Email}\}},\
\underbrace{\textsc{Style}}_{\in\{\text{concise},\ \text{detailed}\}})
\]
with stochastic policy $\pi_\theta(a_t\mid z_t,u_i)$ producing per-head probabilities.

\subsection{Reward Signals and Personalization}\label{sec:reward}
Per-turn utility is a personalized composition
\begin{equation}
R_i(z_t,a_t) \;=\; \alpha_i(z_t)\,R_{\text{user}}(z_t,a_t)
\;+\; \beta_i(z_t)\,R_{\text{tool}}(\tau_t)
\;+\; \gamma_i(z_t)\,R_{\text{eng}}(z_t,a_t),
\label{eq:reward_comp}
\end{equation}
where $\tau_t\in\{\textsf{success},\textsf{failure}\}$ is the logged tool-outcome label (Table~\ref{tab:tool_rubric}), and $R_{\text{tool}}(\varnothing)\!\equiv\!0$ when no tool is invoked. All components are computed from observables in $z_t$ and the chosen action $a_t$; we do not assume access to latent state.

\paragraph{Early-turn information-gain (curiosity).}
To mitigate cold-start and selection-biased feedback, we add a small, early-turn user-model information-gain bonus that rewards reductions in archetype uncertainty:
\begin{equation}
r^{\text{curiosity}}_t \;=\; \max\!\big\{0,\; H\!\big(p_{t-1}(y)\big) - H\!\big(p_t(y)\big)\big\},
\label{eq:curiosity}
\end{equation}
where $y$ is a latent interaction archetype (literacy $\times$ self-efficacy) and $p_t(y)$ is the posterior of a lightweight user model combining dialogue features and survey priors. The intrinsic weight $\lambda_t$ is applied only in the first $K$ turns (we use $\lambda\in\{0.1,0.2\}$; $K{=}2$ in our simulator) and then decays to zero. The total per-turn signal for learning/evaluation is $R_i + \lambda_t\, r^{\text{curiosity}}_t$, with curiosity complementing (never replacing) $R_{\text{user}}$ and $R_{\text{tool}}$.

\subsection{Offline Policy Evaluation (OPE)}
We evaluate counterfactual decision policies over $(\textsc{Tool},\textsc{Style})$ from deployment logs, following logged-bandit practice \citep{swaminathan2015counterfactual,dudik2011doubly,thomas2016data}. We interpret OPE as a per-turn contextual bandit evaluation conditioned on $z_t$, and do not claim unbiased long-horizon policy value.

\paragraph{Behavior-policy reconstruction and calibration.}
For each decision head we fit a probabilistic behavior model $\hat{\pi}_b^{h}(a^h\mid x_t)$ on features $x_t=(z_t,u_i)$ (turn features and user attributes) to approximate logging propensities. We assess calibration via Expected Calibration Error (ECE) per head. For primary reporting we use self-normalized importance sampling (SNIPS)\citep{swaminathan2015self} for the objective component and augmented inverse propensity weighting (AIPW) for satisfaction.

\paragraph{Behavior policy and importance ratios.}
We model the logging policy as head-factorized over $\{\textsc{Tool},\textsc{Style}\}$ and fit
cross-fitted, calibrated classifiers for each head to obtain per-head propensities
$\hat{\pi}_b^{h}(a_t^{h}\mid x_t)$. Assuming head-wise factorization conditional on $x_t$, the joint
logging propensity is $\hat{\pi}_b(a_t\mid x_t)=\prod_{h}\hat{\pi}_b^{h}(a_t^{h}\mid x_t)$, and the
target policy factorizes analogously. For both the SNIPS estimator and the
augmentation term in the AIPW estimator, we use the same joint importance ratio
\[
w_t \;=\; \prod_{h\in\{\textsc{Tool},\textsc{Style}\}}
\frac{\pi^{h}(a^h_t\mid x_t)}{\hat{\pi}_b^{h}(a^h_t\mid x_t)}\,,
\]
with ratio clipping at $c{=}50$ and session-level bootstrap confidence intervals.

\paragraph{Estimators and clipping.}
We clip all importance ratios at $c{=}50$. Our primary estimators are:
\[
\widehat{V}_{\text{SNIPS}}(\pi)=\frac{\sum_t w_t\, r_t}{\sum_t w_t},
\qquad
\widehat{V}_{\text{AIPW}}^{\text{user}}(\pi)
=\frac{1}{T}\sum_t\!\left(\hat q_{\text{user}}(x_t,\pi)+\frac{m_t}{\hat p_{\text{rate},t}}\,w_t\,[r^{\text{user}}_t-\hat q_{\text{user}}(x_t,a_t)]\right),
\]
where $m_t\in\{0,1\}$ indicates whether a rating is observed and $\hat p_{\text{rate},t}$ is a rating-propensity model. 

\paragraph{Total reward composition.}
For interpretability we report an unweighted objective
$R_{\text{obj}} = R_{\text{tool}} + R_{\text{eng}}$ with SNIPS, alongside a
$R_{\text{total}}$ that applies fixed, pre-specified literacy-aware weights to form
$R_{\text{total}} = \alpha_w R_{\text{user}} + \beta_w R_{\text{tool}} + \gamma_w R_{\text{eng}}$.
Here $(\alpha_w,\beta_w,\gamma_w)$ is a deterministic function of literacy strata
(low-literacy: $(0.6,0.2,0.2)$; high-literacy: $(0.3,0.5,0.2)$).
We estimate $\mathbb{E}_\pi[\beta_w R_{\text{tool}} + \gamma_w R_{\text{eng}}]$ with SNIPS and
$\mathbb{E}_\pi[\alpha_w R_{\text{user}}]$ with AIPW using a rating-propensity correction.
Because $R_{\text{obj}}$ (unweighted) and $R_{\text{total}}$ (weighted) use different compositions and estimators
(SNIPS vs.\ AIPW), columns are reported on their native scales and do not sum algebraically in Table \ref{tab:ope_overall}.

\paragraph{Diagnostics.}
Calibration, selection, and coverage diagnostics (rating missingness and propensity AUC, per-head ECE, clipping rate) are reported in Table \ref{tab:ope_diagnostics}.
\begin{table}[t]
\centering
\caption{Off-Policy Evaluation Diagnostics on Pilot Logs}
\label{tab:ope_diagnostics}
\begin{tabularx}{\textwidth}{@{}llX@{}}
\toprule
\textbf{Diagnostic} & \textbf{Value} & \textbf{Notes} \\
\midrule
\addlinespace[0.3em]
\textit{Data Quality} \\
\quad Sessions & 23 & 82.6\% agreement between \texttt{conv\_len} and \texttt{turn\_index}; 1.1\% boundary conflicts \\
\quad Rating rate & 0.80 & Fraction of turns with ratings \\
\addlinespace[0.5em]

\textit{OPE Correction} \\
\quad Rating-propensity AUC & 0.712 & For AIPW importance weighting \\
\quad Clipping rate ($c = 50$) & $\leq$0.29\% & Low importance weight explosion \\
\addlinespace[0.5em]

\textit{Policy Calibration (ECE)} \\
\quad TOOL head & 0.157 & \multirow{2}{=}{\parbox{7cm}{Expected calibration error of behavior policy per prediction head}} \\
\quad STYLE head & 0.050 & \\
\addlinespace[0.5em]

\textit{Tool Performance} \\
\quad Web search & 81.6\% & \multirow{3}{=}{\parbox{7cm}{Success rates by tool type. Top failures: \texttt{no\_url\_provided}, \texttt{no\_visualization}}} \\
\quad Code execution & 80.7\% & \\
\quad Email & 85.7\% & \\
\addlinespace[0.3em]
\bottomrule
\end{tabularx}
\end{table}

\subsection{Simulator with Hidden Archetypes}
To study early exploration and personalization without on-policy trials, we build a lightweight simulator with latent archetypes $y\in\{\text{L}_{\text{low/high}}\times\text{E}_{\text{low/high}}\}$. Each episode samples (i) a health timeseries table (sleep/HRV/steps, etc.) and tasks requiring rolling computations/plots and (ii) wellness APIs (e.g., \texttt{set\_reminder}, \texttt{log\_sleep}) with verifiable end-states. The user-simulator returns a natural reply, a rubric-based rating, and tool outcome labels (Table~\ref{tab:tool_rubric}). We compare policies with/without the curiosity term (Eq.~\ref{eq:curiosity}) on: final return, goal success, pass@3, trait-identification turn/accuracy (from $p_t(y)$), and archetype-aligned action rate. In all simulator runs, the base generator is frozen; policies operate only through the discrete heads. 

\subsection{Reproducibility}
We release de-identified logs of the rated conversational messages, and the code for all evaluations and simulations at \url{https://github.com/stevenshci/NeurIPS-MTI-LLM}.

\section{Pilot Study Results}
We analyzed 350 conversation turns with 280 ratings from our initial pilot with 7 users. While this sample size limits statistical conclusions, we report descriptive patterns that informed our framework design and motivated the new OPE and simulation studies. Ratings showed no meaningful temporal autocorrelation (lag-1: $r{=}0.10$).

\subsection{User Baseline Influence}
Users with above-median self-efficacy averaged $4.3{\pm}0.5$ ratings 
vs.\ $3.9{\pm}0.7$ for below-median. Table~\ref{tab:baseline-effects} shows baseline metrics' influence on interaction behaviors. User archetypes revealed varied satisfaction patterns: low literacy/high efficacy users reported highest median ratings (4.5), while high literacy/low efficacy reported lowest (3.8).

\begin{table}[h]
\centering
\caption{Correlations between baseline metrics and interaction behavior ($N{=}7$). Values are descriptive.}
\label{tab:baseline-effects}
\begin{tabular}{lc}
\toprule
Baseline Metric & Correlation ($r$) \\
\midrule
Health literacy vs.\ tool usage        & 0.302 \\
Self-efficacy vs.\ conversation length & $-0.441$ \\
\bottomrule
\end{tabular}
\end{table}

\subsection{Tool Use Impact}
Tool usage showed descriptive differences: tool-invoked responses averaged 4.02 vs.\ 4.38 for non-tool responses (Cohen's $d{=}{-}0.40$). Table~\ref{tab:tool-impact} details category-specific patterns, with a success--failure gap of $+0.50$. Tool responses showed $1.40{\times}$ higher variance ($\sigma^2{=}0.94$ vs.\ $0.67$), reflecting high-risk/high-reward dynamics.

\begin{table}[h]
\centering
\caption{Impact of tool use on ratings (1-5) by category. (Email is n=7 turns, 1 failure).}
\label{tab:tool-impact}
\begin{tabular}{lccc}
\toprule
Tool Category & Success Mean & Failure Mean& Gap \\
\midrule
Web Searcher       & 4.01  & 3.53 & +0.48  \\
Code Executor      & 4.09  & 3.33 & +0.76  \\
Email     & 5.00  & 5.00 & 0.00   \\
\midrule
All Tools          & 4.08 & 3.58 & +0.50  \\
\bottomrule
\end{tabular}
\end{table}

Mixed-effects modeling suggested minimal between-user heterogeneity in unconditional ratings: only $1.6\%$ of variance was attributable to user identity (ICC=$0.016$). This does not preclude \emph{context-conditioned} heterogeneity: as shown below, per-archetype differences under counterfactual policies are large.

\subsection{Implicit Feedback and Performance Degradation}
Response time negatively correlated with ratings ($r{=}{-}0.18$): $<10$s responses averaged 4.25 vs.\ 3.98 for $>30$s. Message length showed a small positive association ($r{=}0.07$); high-rated messages (4--5) averaged 2{,}123 characters vs.\ 1{,}275 for low-rated (1--2). Ratings declined from 4.36 (turns 1--5) to 4.12 (15+), while tool usage peaked at 70\% (turns 5--10) then dropped to 26.3\% for turns 15+. Latency peaked mid-conversation (25s at turns 6--10), and message length decreased 3.4\% from early to late turns.

\subsection{Counterfactual Policy Analysis}
Using SNIPS and AIPW with session-level bootstrap CIs, we compared four policies over discrete decision heads: \textsc{NoTool} (never invoke a tool), \textsc{AlwaysTool} (always invoke a tool; tool type drawn from the logged conditional distribution to maintain overlap), \textsc{HeuristicGated} (simple rule-based triggers), and \textsc{PersonalizedWeights}(same heuristics with a literacy-aware bias that raises or lowers the gate). On average, \textsc{AlwaysTool} attained the highest $R_{\text{total}}$ on our logs (0.304, 95\% CI [0.001, 0.524]). However, per-archetype slices revealed large and structured heterogeneity: heavy tool use benefited high-literacy/low-efficacy users across both objective and satisfaction components, but harmed low-literacy/high-efficacy users on both (e.g., $\Delta$Objective $-0.315$; $\Delta$Satisfaction $-1.436$ vs.\ \textsc{NoTool}). Tables~\ref{tab:ope_overall}--\ref{tab:ope_archetype} summarize results.

\begin{table}[t]
\centering
\caption{Offline policy evaluation on deployment logs. 
$R_{\text{obj}}$: SNIPS (unweighted objective); $R_{\text{user}}$: AIPW (z-score).
$R_{\text{total}}$: literacy-weighted combination. Brackets show session-level bootstrap 95\% CIs for $R_{\text{total}}$.}
\label{tab:ope_overall}
\begin{tabular}{lccc}
\toprule
Policy & $R_{\text{obj}}$ (SNIPS) & $R_{\text{user}}$ (AIPW) & $R_{\text{total}}$ \\
\midrule
\textsc{NoTool}              & 0.328 & $-0.623$ & \textbf{0.044} \;[{-}0.045,\; 0.198] \\
\textsc{AlwaysTool}          & 0.229 & $-0.654$ & \textbf{0.304} \;[\textbf{0.001},\; 0.524] \\
\textsc{HeuristicGated}      & 0.309 & $-0.625$ & \textbf{0.006} \;[{-}0.111,\; 0.174] \\
\textsc{PersonalizedWeights} & 0.253 & $-0.656$ & \textbf{0.113} \;[{-}0.016,\; 0.284] \\
\bottomrule
\end{tabular}
\end{table}

\begin{table}[t]
\centering
\caption{Heterogeneity by archetype: difference in OPE estimates (AlwaysTool $-$ NoTool). Positive is better.}
\label{tab:ope_archetype}
\begin{tabular}{lrr}
\toprule
Archetype & $\Delta$ Objective & $\Delta$ Satisfaction \\
\midrule
\(\mathrm{L_{high}\times E_{high}}\) & +0.575 & $-0.107$ \\
\(\mathrm{L_{high}\times E_{low}}\)  & +0.595 & \textbf{+0.525} \\
\(\mathrm{L_{low}\times E_{low}}\)   & +0.165 & $-0.431$ \\
\(\mathrm{L_{low}\times E_{high}}\)  & \textbf{$-0.315$} & \textbf{$-1.436$} \\
\bottomrule
\end{tabular}
\end{table}

\subsection{Simulation with Curiosity}
We built a lightweight simulator with hidden archetypes (literacy $\times$ self-efficacy) and verifiable tool tasks. Adding a small early information-gain bonus for the first two turns improved goal success (0.935 $\rightarrow$ 0.970), pass@3 (0.95 $\rightarrow$ 0.98), final return ($-3.162 \rightarrow -2.329$), and reduced trait-identification time (6.41 $\rightarrow$ $\approx$5.7 turns), with a small expected dip in archetype-aligned actions due to early probing (Table~\ref{tab:sim_curiosity}).
\begin{table}[t]
\centering
\caption{Simulation with hidden archetypes; $\lambda$ controls the early information-gain bonus (applied in the first two turns). Trait-ID turn: lower is better.}
\label{tab:sim_curiosity}
\begin{tabularx}{\textwidth}{lXXXXXX}
\toprule
Policy & Final Return & Goal Succ. & pass@3 & Trait-ID Turn & Trait Acc. & Arch. Align \\
\midrule
\textsc{Heuristic}                 & $-2.908$ & 0.515 & 0.505 & 6.315 & 0.050 & 0.503 \\
\textsc{Personalized}              & $-3.162$ & \textbf{0.935} & \textbf{0.950} & 6.415 & 0.050 & 0.424 \\
\textsc{Pers+Curiosity} ($\lambda{=}0.10$) & \textbf{$-2.401$} & \textbf{0.965} & \textbf{0.975} & \textbf{5.655} & 0.075 & 0.412 \\
\textsc{Pers+Curiosity} ($\lambda{=}0.20$) & \textbf{$-2.329$} & \textbf{0.970} & \textbf{0.980} & 5.860 & 0.045 & 0.410 \\
\bottomrule
\end{tabularx}
\vspace{0.25em}
\textsc{Heuristic} (mirror of \textsc{HeuristicGated}; \textsc{Style} defaults to \emph{concise} unless the user asks to "explain/why/how"), \textsc{Personalized} (mirror of \textsc{PersonalizedWeights} with \textsc{Style} also adapted by archetype).
\end{table}

\paragraph{Summary.}
The pilot study suggests three findings: (i) tool use is \emph{high-variance} with typed, fixable failure modes; (ii) average trends mask \emph{large, structured heterogeneity} by archetype; and (iii) performance degrades over turns, consistent with a need for targeted early information gathering and personalization. The OPE and simulator results build on these observations and provide preliminary validation for our evaluation-first, personalization-first framework.

\section{Implications for Future RL Frameworks}\label{sec:rl_implications}

We outline design implications and a blueprint for RL-style personalization informed by our OPE and simulator findings. We do not train an end-to-end RL policy in this work, and the base generator remains frozen. OPE indicates that a uniform heavy-tool policy can lift averages while harming specific subgroups (Table~\ref{tab:ope_archetype}), and simulation suggests a brief early information-gain phase improves task success and shortens trait identification (Table~\ref{tab:sim_curiosity}). Motivated by these observations, we frame coaching as a user-conditioned POMDP with a belief state $z_t=f_\phi(h_t,u_i,m_t)$ and a policy acting over discrete decision heads (\textsc{Tool}/\textsc{Style}). As a hypothesis for future work, we propose archetype-aware weighting that increases $\beta$ for $\mathrm{L_{high}\!\times E_{low}}$ users, reduces $\beta$ and raises $\alpha$ for $\mathrm{L_{low}\!\times E_{high}}$, and pairs tools with explanatory scaffolds for $\mathrm{L_{low}\!\times E_{low}}$; a short, bounded curiosity term (Eq.~\ref{eq:curiosity}) can accelerate preference elicitation in the first turns.

Moreover, we recommend an offline-first loop: fit or select population head policies from logs via contextual-bandit objectives while keeping the generator frozen; initialize user-specific weights from cluster priors (eHEALS/GSE strata and early interaction features) and adapt with strong regularization; and gate deployments by OPE diagnostics. End-to-end policy learning and prospective validation are left to future work.

\section{Conclusion}
We studied a deployed, tool-augmented LLM health coach with real users and extended our observations with two components: (i) offline policy evaluation (OPE) on deployment logs to compare counterfactual decision policies, and (ii) a lightweight simulator with hidden archetypes to test brief early information-gain phases. Three findings emerge. First, while uniform heavy tool usage maximizes average reward on logs, OPE reveals large heterogeneity across archetypes: the same policy benefits high-literacy/low-efficacy users but harms low-literacy/high-efficacy users on both objective and satisfaction measures. Second, adding a small, bounded early information-gain bonus consistently improves simulated task success, pass@3, and reduces trait-identification time, supporting a “probe-then-personalize” strategy. Third, the OPE pipeline (behavior-policy reconstruction, calibration diagnostics, clipping, session-bootstrap CIs, and rating-selection correction) is stable enough-per diagnostics, to guide offline-first policy iteration without new user trials.

These results provide an operational path forward: freeze the generator, learn subgroup-aware decision heads (instantiated here as literacy-aware) using typed rewards (objective outcomes + satisfaction), and use OPE to compare policies offline. Critically, report per-archetype metrics to catch policies that improve average performance while harming specific user groups.

\paragraph{Limitations and next steps.}
Our deployment sample is small and demographically narrow, limiting generalizability. Behavior propensities are reconstructed rather than logged, and Tool-head calibration is moderate, which can bias IPS/SNIPS; we mitigate with AIPW, clipping, and diagnostics (ECE). Our current evaluation weights and personalized routing are conditioned on literacy only; incorporating self-efficacy into weighting and head policies is a planned extension.

\bibliographystyle{unsrtnat} 

\end{document}